\documentclass[10pt,twocolumn,letterpaper]{article}

\usepackage{wacv}              % To produce the CAMERA-READY version
\usepackage{graphicx}
\usepackage{amsmath}
\usepackage{amssymb}
\usepackage{booktabs}
\usepackage{times}
\usepackage{microtype}
\usepackage{epsfig}
\usepackage[table,xcdraw,dvipsnames]{xcolor}
\usepackage{caption}
\usepackage{float}
\usepackage{placeins}
\usepackage{color, colortbl}
\usepackage{stfloats}
\usepackage{enumitem}
\usepackage{tabularx}
\usepackage{xstring}
\usepackage{multirow}
\usepackage{xspace}
\usepackage{url}
\usepackage{algorithm}
\usepackage{algpseudocode}
\usepackage[frozencache,cachedir=.]{minted}
\usepackage{pifont}
\usepackage[accsupp]{axessibility}  % Improves PDF readability for those with disabilities.

\usepackage[pagebackref,breaklinks,colorlinks]{hyperref}

\usepackage[capitalize]{cleveref}
\crefname{section}{Sec.}{Secs.}
\Crefname{section}{Section}{Sections}
\Crefname{table}{Table}{Tables}
\crefname{table}{Tab.}{Tabs.}
\creflabelformat{equation}{#2#1#3}

\newif\ifDEBUG
\DEBUGfalse

\newif\ifSUPPLEMENTAL
\SUPPLEMENTALtrue
\ifDEBUG
    \newcommand{\new}[1]{\textcolor{blue}{#1}}
    \newcommand{\NJE}[1]{\textcolor{red}{[NJE: #1]}}
    \newcommand{\PJ}[1]{\textcolor{blue}{[PJ: #1]}}
    \newcommand{\JD}[1]{\textcolor{purple}{[JD: #1]}}
    \newcommand{\YHL}[1]{\textcolor{olive}{[YHL: #1]}}
    \newcommand{\GKT}[1]{\textcolor{violet}{[GKT: #1]}}
    \newcommand{\GL}[1]{\textcolor{red}{[GL: #1]}}
    \newcommand{\GR}[1]{\textcolor{red}{[GR: #1]}}

\else
    \newcommand{\new}[1]{#1}
    \newcommand{\NJE}[1]{}
    \newcommand{\PJ}[1]{}
    \newcommand{\JD}[1]{}
    \newcommand{\YHL}[1]{}
    \newcommand{\GKT}[1]{}
    \newcommand{\GL}[1]{}
    \newcommand{\GR}[1]{}
\fi

\newcommand\UAt{U$_{\text{A}}(n)$}
\newcommand\ULt{U$_{\text{L}}(n)$}

\newcommand\MPre{$M$}

\def\wacvPaperID{1151}
\def\confName{WACV}
\def\confYear{2025}

\begin{document}

%\title{Token Pruning using Latency and Workload Size Disparities for ViTs on the Edge}
\title{Pruning One More Token is Enough: Leveraging Latency-Workload Non-Linearities for Vision Transformers on the Edge}
\author{Nick John Eliopoulos$^1$
\and
Purvish Jajal$^1$
\and
James C. Davis$^1$
\and
Gaowen Liu$^2$
\and
George K. Thiravathukal$^3$
\and
Yung-Hsiang Lu$^1$
}

\maketitle

%%%
%%% Abstract
%%%
\begin{abstract}
This paper investigates how to efficiently deploy vision transformers on edge devices for small workloads. 
Recent methods reduce the latency of transformer neural networks by removing or merging tokens, with small accuracy degradation.
However, these methods are not designed with edge device deployment in mind: they do not leverage information about the latency-workload trends to improve efficiency.
We address this shortcoming in our work.
First, we identify factors that affect ViT latency-workload relationships.
Second, we determine token pruning schedule by leveraging non-linear latency-workload relationships.
Third, we demonstrate a training-free, token pruning method utilizing this schedule. 
We show other methods may increase latency by 2-30\%, while we reduce latency by 9-26\%.
For similar latency (within 5.2\% or 7ms) across devices we achieve 78.6\%-84.5\% ImageNet1K classification accuracy, while the state-of-the-art, Token Merging, achieves 45.8\%-85.4\%.
\end{abstract}

%%%
%%% Introduction
%%%
\section{Introduction}\label{sec:intro}

%%
% Context
%%
In the past decade, Internet of Things (IoT) and commodity edge devices have become ubiquitous~\cite{cao_overview_2020, lyytinen2004surfing, weiser1999computer}.
Edge devices have become sufficiently powerful, and model miniaturization techniques sufficiently capable, that machine learning (ML) models can be deployed to the network edge for various tasks, including computer vision applications~\cite{goel_survey_2020, lu_textbook_efficient_2022}.
However, state of the art performance on various computer vision tasks is often claimed by large vision transformer (ViT) based neural network~\cite{dosovitskiy_image_2020} architectures.
ViT models such as DeiT~\cite{touvron_deit_2022} and DINOv2~\cite{oquab_dinov2_2023} are not designed nor miniaturized for edge deployment.
Additionally, latency \new{(the time required to do a forward pass given a batch of inputs)} is often of critical importance on edge devices, and only small inputs or workloads can be processed \cite{edge_optim_batch_2022}.

%%%
%%% Prior work
%%%
Prior work has shown that ViTs have high redundancy that can be exploited for latency reduction benefits~\cite{naseer_intriguing_2021}.
One approach involves identifying and removing low-information tokens; this is called token sparsification.
Training-free token sparsification methods such as Token Merging (ToMe) \cite{bolya_token_nodate} have been effective at reducing latency of pre-trained models.
Other approaches like DynamicViT \cite{rao_dynamicvit_2021} can yield better accuracy than training-free methods, but require training on server-grade hardware.

%%%
%%% Our contribution
%%%

%%%
%%% Tail Effect Batch Size Ablation Plot with deit-large 
%%%
\begin{figure*}[t!]
    \setlength{\belowcaptionskip}{-6pt}
    \centering
    \includegraphics[width=0.70\linewidth]{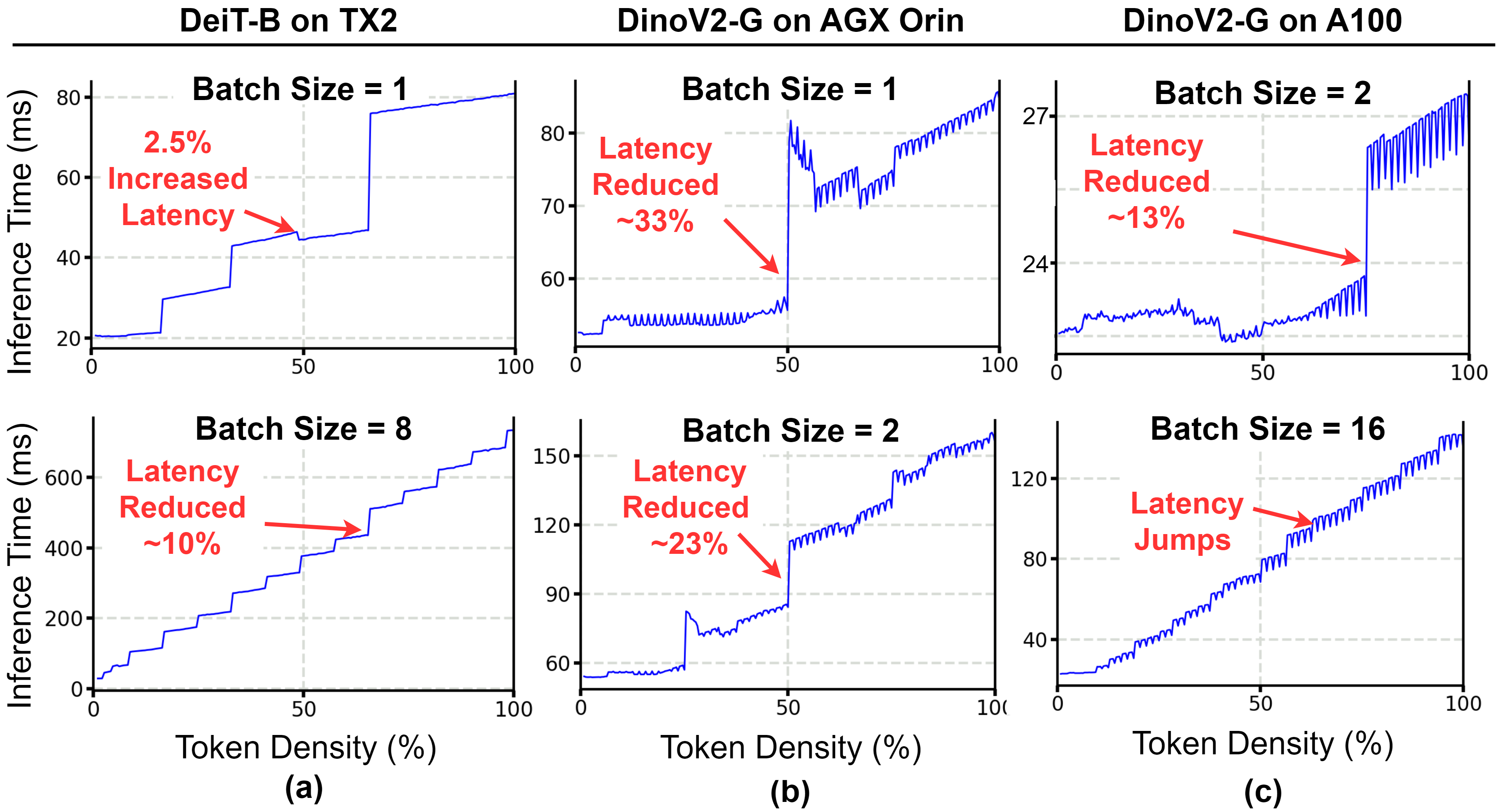}
    %\vspace{-1em}
    \caption{
    Forward pass latency for widely used DeiT-B ($d=768$) and DinoV2-G $(d=1536$) models across various hardware (\cref{tab:hardware_info}) evaluated on the ImageNet1K~\cite{deng_imagenet_2009} classification dataset.
    These plots demonstrate the \textit{variable} and \textit{non-linear} relationship between workload size (as defined in \cref{sec:number_to_prune}) and latency, across a variety of hardware. 
    Consequently, in many cases it is possible to achieve large latency reductions without removing too many tokens.
    This work shows when and how to remove tokens to take advantage of these latency non-linearities.
    }
    \label{fig:tail_effect_batch_ablate}
\end{figure*}

We address three key shortcomings in existing techniques.
(1) Many existing efficient methods do not consider fine-grained hardware or latency characteristics \cite{bolya_token_nodate, liang_not_2022, wei_joint_2023}.
~\cref{fig:tail_effect_batch_ablate} demonstrates the diversity of latency-workload trends across devices and workload sizes.
As a result, there is room to improve efficient methods in a hardware-aware manner by considering this relationship.
(2) Some existing efficient methods may require extensive training \cite{zhou_sp-vit_2022, rao_dynamicvit_2021, chang_making_2023}, hindering the deployment of pre-trained models on edge devices.
(3) Prior work has investigated hardware-aware methods for CNNs \cite{zhang_locp_2023, yu_towards_2020}, but there is little work that shows how to handle or leverage latency-workload behavior for ViTs. 
Finally, these works lack direct measurements of underlying GPU or kernel behavior.
Thus, our work focuses on reducing ViT latency by considering ViT latency-workload relationships, and without requiring training.
\textbf{This paper has the following contributions:}%
\begin{enumerate}
    \item We identify and profile factors that can affect ViT latency-workload relationships.
    \item We propose the first method using latency-workload relationships for deciding ViT token pruning schedules. 
    \item We design a novel training-free token pruning mechanism. For similar latency across various hardware and workload sizes, we achieve 0.46 to 43.7 percentage points higher accuracy than ToMe, a state-of-the-art method.
\end{enumerate}

%%%
%%% Related
%%%
\section{Background and Related Work}
In this section, we review related work on model acceleration and efficient methods for vision transformers (ViT).
Some post-processing or fine-tuning methods such as quantization \cite{liu2021post} are compatible with token sparsification techniques such as our method.

\subsection{Model Acceleration of Vision Transformers}\label{bg:prune}
ViT \cite{dosovitskiy_image_2020} architectures such as DINOv2~\cite{oquab_dinov2_2023} have achieved state-of-the-art accuracy on multiple computer vision tasks, including image classification and object detection.
State-of-the-art models such as DINOv2-G \cite{oquab_dinov2_2023} have over 1 billion parameters.
It is important to address the efficiency of ViT models when deploying on edge devices.
Numerous techniques exist for accelerating ViT models, including: quantization \cite{liu2021post}, knowledge distillation \cite{wu2022tinyvit}, low-rank factorization \cite{chen2021scatterbrain}, and optimized kernels for attention \cite{dao_flashattention_2022}.
In general, these techniques either remove redundant information or reduce network layer compute.

One approach related to redundant information removal is token sparsification~\cite{bolya_token_nodate, kong_spvit_2022, renggli_learning_2022, rao_dynamicvit_2021}.
These methods are easily applied to a variety of ViT architectures~\cite{touvron_deit_2022, oquab_dinov2_2023, dosovitskiy_image_2020}.
One advantage of sparsification methods is that they often do not necessarily require training or fine-tuning~\cite{wang_zero-tprune_2023, bolya_token_nodate}.
However, some methods \cite{rao_dynamicvit_2021, zhou_sp-vit_2022, chang_making_2023, wei_joint_2023} may require significant training time, \eg 100+ epochs, to recover more accuracy and yield latency reductions.
Training poses a high barrier to application; training-free methods are more accessible \cite{bolya_token_nodate, wang_zero-tprune_2023}.

\subsection{Latency-Workload Relationship}\label{bg:gpu_tail}
The premise behind token pruning is that reducing the workload (tokens) can decrease latency.
However, this relationship can be non-linear, as demonstrated in \cref{fig:tail_effect_batch_ablate}.
This relationship can stem from how workloads are dispatched to the GPU by an ML framework, framework overhead~\cite{framework_tax_2023}, and kernel design decisions~\cite{nvidia_cuda_practices_guide}.
~\cref{tab:cuda_causes} illustrates primary causes of kernel inefficiency.
%%%
%%% Qualitative eval table
%%%
\begin{table}[b!]
    \centering
    %\setlength{\belowcaptionskip}{-16pt}
    %\setlength{\tabcolsep}{2pt}
    %%%
    %%% Pruning Method Info
    %%%
    \begin{tabular}{lll}
        \toprule
        \# & \textbf{Cause} & \textbf{Diagnostic Metrics}\\
        \midrule
        1 & Launch Config & Occupancy, Grid Size\\ 
        2 & Memory Usage & Memory Bandwidth\\ 
        3 & Instruction Usage & Math Pipe Stalls, Instructions\\ 
        \bottomrule
    \end{tabular}
    %%%
    %%% Table Caption
    %%%
    \caption{
    Primary causes for kernel inefficiency, with associated metrics \cite{nvidia_cuda_practices_guide, pmpp_2023}. 
    Note these causes can co-occur.
    }
    \label{tab:cuda_causes}
\end{table}

An example of Cause 1 occurs where a kernel grid size is chosen such that a partial wave of computation must be launched on the GPU --- this can lead to a phenomenon termed the GPU tail effect \cite{nvidia_gpu_perf}.
A partially filled wave incurs the same latency cost as a fully filled one, and this effect can compound across layers.
Thus, even minor adjustments in workload size can result in significant latency changes due to the cumulative overhead of partial waves across layers.
For example, on the NVIDIA AGX Orin \cite{nvidia_agxorin}, removing one token (97 to 96) can decrease latency by up to 33\% (\cref{fig:tail_effect_batch_ablate}.b).

There are many factors that affect latency.
Differences in hardware, changes across ML framework versions, and reliance on proprietary backend libraries like cudNN~\cite{nvidia_cudnn_2014} and cuBLAS~\cite{nvidia_cublas} complicate the modeling and prediction of neural network latency.
To illustrate this difficulty, we show more examples in~\cref{sec:measure_vs_predict}.

Prior work has exploited the tail effect, which is related to Cause 1, to guide pruning methods for convolutional networks \cite{zhang_locp_2023, yu_towards_2020}.
Kernel-based optimization methods such as FlashAttention \cite{dao_flashattention_2022} may attempt to choose kernel launch configurations that maximize occupancy.
Quantization addresses Cause 2 by employing data types with fewer bits than 32-bit floating point for network parameters.
Cause 3 can be addressed by choosing low-level operators that might be faster at the cost of precision, or vice-versa \cite{nvidia_cuda_practices_guide}.

In this work, we address token pruning in the context of ViT models.
Previous work frames CNN channel pruning in the context of Cause-1 problems, specifically the GPU tail effect.
However, ViT token pruning mechanisms are fundamentally different from previous CNN channel pruning \cite{cnn_channel_prune_2017} approaches due to architectural differences between CNNs and ViTs.
We hypothesize and later demonstrate that latency-workload relationships can also be leveraged to make better token pruning decisions for ViTs.

%%%
%%% Methods: Schedule Selection
%%%
\section{Token Pruning with Consideration of Latency and Workload Size}
In \cref{bg:prune} and \cref{bg:gpu_tail}, we discussed the advantages of training-free token pruning methods and how previous work considered latency and workload size relationships for efficiency benefits.
Therefore, we set two design goals:
(1) require no training or fine-tuning of a pre-trained ViT model;
and
(2) achieve better accuracy-latency tradeoffs by pruning tokens according to these relationships.
\cref{tab:qualitative_eval} illustrates qualitative differences between our work and others as a result of these goals.

%%%
%%% Qualitative Difference Table
%%%
%%%
%%% Qualitative eval table
%%%
\begin{table}[b!]
    \centering
    %\setlength{\belowcaptionskip}{-4pt}
    %\setlength{\tabcolsep}{8pt}
    %%%
    %%% Pruning Method Info
    %%%
    \begin{tabular}{p{0.30\linewidth}cc}
        \toprule
        \textbf{Method} & \textbf{Training Free} & \textbf{Hardware Aware}\\
        %& ~ & \textbf{Aware?}\\
        \toprule
        Ours & \textcolor{ForestGreen}{\ding{51}} & \textcolor{ForestGreen}{\ding{51}}\\
        \midrule
        %\midrule
        ToMe \cite{bolya_token_nodate} & \textcolor{ForestGreen}{\textcolor{ForestGreen}{\ding{51}}} & \textcolor{Maroon}{\ding{55}}\\
        %Zero-TPrune \cite{wang_zero-tprune_2023} & \textcolor{ForestGreen}{\textcolor{ForestGreen}{\ding{51}}} & \textcolor{Maroon}{\ding{55}}\\
        EViT \cite{liang_not_2022} & \textcolor{ForestGreen}{\textcolor{ForestGreen}{\ding{51}}} & \textcolor{Maroon}{\ding{55}}\\
        Top-K \cite{which_tokens_to_use_2023},\cite{liang_not_2022} & \textcolor{ForestGreen}{\textcolor{ForestGreen}{\ding{51}}} & \textcolor{Maroon}{\ding{55}}\\
        DynamicViT \cite{rao_dynamicvit_2021} & \textcolor{Maroon}{\ding{55}} & \textcolor{Maroon}{\ding{55}}\\
        %STViT \cite{chang_making_2023} & \textcolor{Maroon}{\ding{55}} & \textcolor{Maroon}{\ding{55}}\\
        TPS \cite{wei_joint_2023} & \textcolor{Maroon}{\ding{55}} & \textcolor{Maroon}{\ding{55}}\\
        \bottomrule
    \end{tabular}
    %%%
    %%% Table Caption
    %%%
    \caption{
    Qualitative differences between our pruning approach and similar techniques. 
    We consider ViT latency-workload relationships to guide our token pruning.
    %, making us hardware aware. 
    \GKT{I feel like "making us hardware aware) is a bit weak. Aren't we really saying (based on the evidence in our table) that what makes our approach unique is that we incorporate hardware awareness? Will it hurt us to say anything more specific to make the caption more self-contained?}
    \NJE{Addressed.}
    }
    \label{tab:qualitative_eval}
\end{table}

\begin{figure*}[t!]
    \centering
    \setlength{\belowcaptionskip}{-8pt}
    \includegraphics[width=0.80\linewidth]{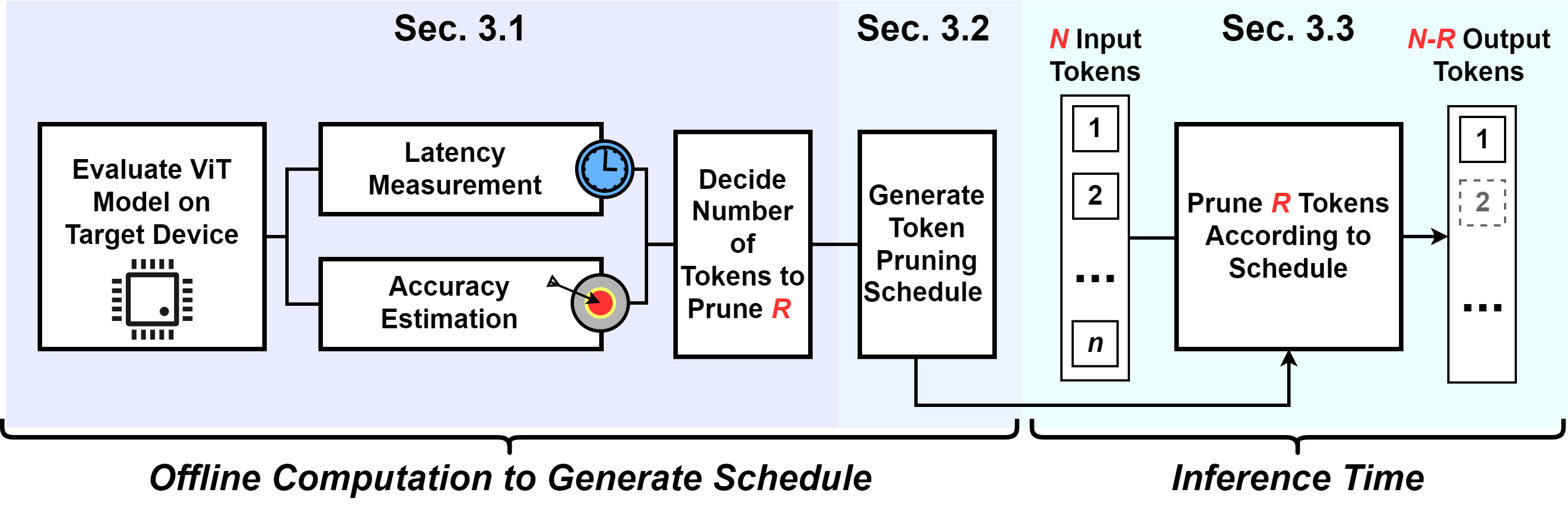}
    \caption{
    Illustration of our method to decide a pruning schedule (left) and how we prune according to the schedule at inference time (right), which are discussed in the sections shown at the top of the illustration.
    }
    \label{fig:overall_flowchart}
\end{figure*}

As depicted in \cref{fig:overall_flowchart}, our token pruning approach consists of two main parts.
First, we establish a pruning schedule for specific model-device pairs, which involves determining the number of tokens to prune and the layers where pruning occurs.
Second, we devise a training-free technique for pruning non-informative tokens at inference time.
In \cref{sec:number_to_prune} we show how to decide the number of tokens to prune based on the latency-workload relationship.
Next, in \cref{sec:schedule} we explain our choice of which layers to prune, completing our offline pruning schedule selection.
\cref{sec:pruning} describes our token pruning mechanism which is used at inference time.
\new{Last, \cref{sec:qual} clarifies qualitative differences between our method and existing approaches.}

%%%
%%% Latency-Workload Figure
%%%
\begin{figure}[b!]
    \centering
    \setlength{\belowcaptionskip}{-4pt}
    \includegraphics[width=0.80\linewidth]{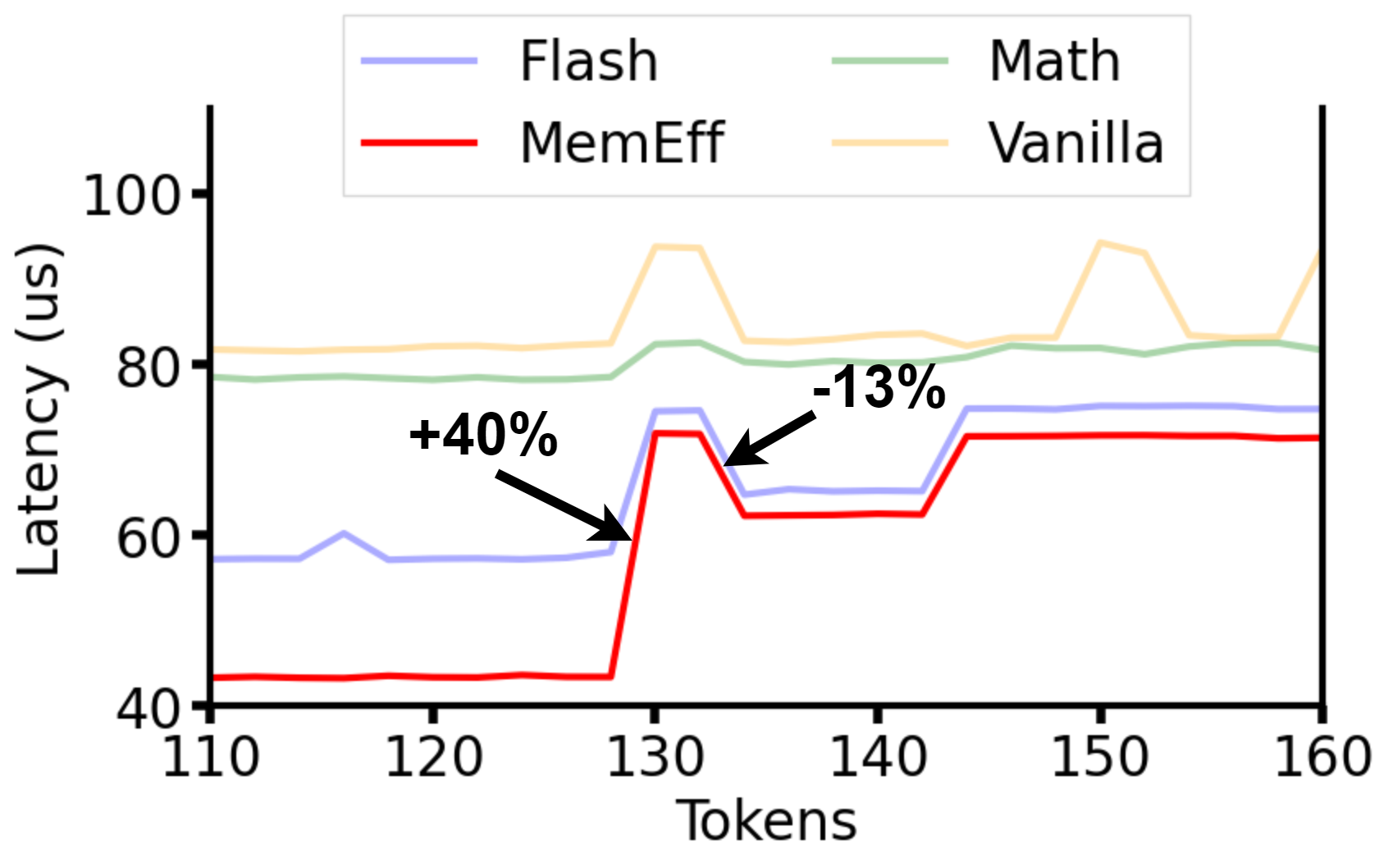}
    \caption{
    Latency-workload characteristics of attention operators in PyTorch \cite{torch_sdpa_2024}.
    Flash, MemEfficient, and Math are optimized, while Vanilla is not.
    Median latency was measured over 100 runs for each token count. 
    All measurements had an IQR $<$ 1$\mu$s. \GKT{Not sure what 1us is here; do you mean microseconds? That should be Greek Mu, if so.}
    The two annotated latency changes of \texttt{MemEff} are discussed in \cref{sec:number_to_prune}.
    }
    \label{fig:torch_kernels}
\end{figure}

%%%
%%% Overall method flowchart figure
%%%

%%%
%%% Pruning Schedule Choice
%%%
\subsection{Deciding a Number of Tokens to Prune for our Pruning Schedule}\label{sec:number_to_prune}
\noindent\textbf{ViT workload size.}
Before explaining our method we discuss the core operator of ViT models, the attention mechanism \cite{vaswani_attention_2017}.
ViT models are partially parameterized by an embedding dimension $d$, with inputs characterized by a batch size $b$ and a number of tokens $n$.
The input size, or workload, for attention is a tensor with shape $b \times n \times d$.
\cref{fig:tail_effect_batch_ablate} demonstrates how varying $b$, $n$, and $d$ affects the relationship between workload size and latency across various devices.

This wide variety of behavior across devices and workload sizes leads us to consider how ViT latency-workload relationships can be reasonably measured or modeled.
We now explain underlying reasons for latency-workload non-linearity, and how we decide to measure it.\\

\noindent\textbf{Measuring latency behavior effects vs. predicting them.}\label{sec:measure_vs_predict}
As previously mentioned, the GPU tail effect is one phenomenon that arises due to suboptimal kernel grid size choice.
Prior work has modeled \cite{liu_latency-aware_2021} or utilized \cite{zhang_locp_2023, yu_towards_2020} GPU tail effect behavior with respect to convolutional neural networks CNNs.
Other work has noted the (sometimes drastic) effect of latency overhead from ML frameworks~\cite{framework_tax_2023}.

Yu et al. \cite{yu_towards_2020} claims that the latency of CNNs increases in discrete jumps, and in all other workload size intervals latency remains the same.
In our experiments, \textit{we find that internal ML framework operator selection can lead to a range of latency behavior. 
This claim for CNNs does not hold in our evaluations for ViTs.}
\cref{fig:torch_kernels} depicts the latency-workload characteristics of various attention operators in PyTorch~\cite{paszke_pytorch_2019} with $b$=1 and varying $n$ for a DinoV2-G~\cite{oquab_dinov2_2023} attention operation.
This was measured on an NVIDIA RTX 3090 Ti GPU.
Additionally, we \textit{find that the tail effect is not always the primary factor 
for drastic latency changes, as was described in previous work.}
We illustrate these findings with measurements of metrics mentioned in~\cref{bg:gpu_tail} and their co-occurrence with latency behavior.  

In one example from \cref{fig:torch_kernels}, the \texttt{MemEff} attention operator features a $\sim$40\% latency increase from 128 to 130 tokens that is correlated with a $\sim$40\% increase in pipeline stall or wait time (Cause 3).
But, the kernel grid size remained the same, indicating that the tail effect (Cause 1) was not the primary cause of this latency increase.
A second example is the  $\sim$13\% latency decrease of \texttt{MemEff} from 132 to 134 tokens is correlated with a 125\% increase in kernel grid size (Cause 1) and a 34\% decrease in pipeline stall time (Cause 2).
These effects seem to be downstream of a different kernel being chosen due to internal heuristics.
Importantly, we note \textit{latency can even decrease as workload increases, due to underlying kernel, hardware, and framework behaviors.}

Considering these observations, we empirically measure the latency-workload relationship due to the difficulty in predicting or modeling its behavior.
The next section describes how we perform this measurement.\\

%%% Frame the problem of token selection
\noindent\textbf{Ranking token importance.}
Given a model \MPre{} with $N$ input tokens, we define the selection of $R$ tokens to prune as a multi-objective optimization problem, balancing latency gains against accuracy degradation.
This is the \textit{offline computation} for selecting a pruning schedule for \MPre{} on a target device, as seen in \cref{fig:overall_flowchart}.

First, we measure latency $L(n)$ of \MPre{} on the target device for each number of tokens to keep $n \in [1, 2, ..., N]$.
Measuring latency and accuracy of \MPre{} is performed with a grid search across $n$ --- \new{this is demonstrated in \cref{alg:offline_lat}.}
\new{Running times for all configurations we evaluate over are listed later in \cref{sec:offline_time}.} 

%%%
%%% Latency Measurement
%%%
\begin{algorithm}
\begin{minted}[fontsize=\footnotesize]{python}
def measure_latency(model, b, N):
    """ b refers to batch size,
    N is the number of tokens 'model' expects """
    #L(n), as a dictionary
    L = {}
    for n in range(1, N):
        # Latency is independent of random inputs
        # model.d is the embedding size (Sec. 3.1)
        x = torch.rand(b,n,model.d)
        # Benchmark for fixed time
        latency = bench(model, x)
        L[n] = latency
    return L
\end{minted}
\caption{Offline Workload Latency Measurement}
\label{alg:offline_lat}
\end{algorithm}%
%%%
%%% Accuracy Measurement
%%%
\begin{algorithm}
    \begin{minted}[fontsize=\footnotesize]{python}
def measure_accuracy(model, dataset, N):
    """ dataset is the eval split,
    N is the number of tokens 'model' expects """
    #A(n), as a dictionary
    A = {}
    for n in range(1, N):
        # Running accuracy for n
        n_acc = 0.0
        for image,label in dataset:
            # Shape (b, N, model.d)
            x = model.embed(image)
            # Shape (b, n, model.d)
            x = random_prune(x, n)
            y_pred = model.predict(x)
            n_acc += sum(y_targ == label)
        A[n] = n_acc / len(dataset)
    return A
\end{minted}
\caption{Offline Accuracy Degradation Estimate}
\label{alg:offline_acc}
\end{algorithm}

Second, we estimate the accuracy $A(n)$ of \MPre{} after pruning tokens.
We need to measure $A(n)$ in a way that does not depend on our pruning schedule selection, however.
Furthermore it is useful to underestimate the accuracy of \MPre{} so our selection algorithm is hesitant to remove too many tokens, which can degrade accuracy significantly.
A simple proxy for the accuracy of each $n$ is to apply random token removal after the first layer on \MPre{} \cite{iared_2021}, which we refer to as \texttt{random\_prune}.
\new{\cref{alg:offline_acc} depicts how to compute $A(n)$.}
It is assumed that any token pruning method should be better than random token removal since random token pruning does not consider token information content at all \cite{wei_joint_2023}.
Furthermore, pruning at the first layer will degrade accuracy more than pruning later in the network \cite{bonnaerens_learned_2023}.
Thus, random token pruning is a suitable choice for estimating accuracy.

%% Original - Eqn. Version
Third, in order to solve the multi-objective optimization problem, we need to transform our measurements $L(n)$ and $A(n)$ into utility functions \ULt{} and \UAt{}. 
We want \ULt{} to be normalized to [0,1], and the $n_i$ with minimum latency has $\text{U}_{\text{L}}(n_i)=1$, and the $n_j$ with maximum latency receives $\text{U}_{\text{L}}(n_j)=0$. 
Similarly, $\text{U}_{\text{A}}(n_k)=1$ for $n_k$ with maximum accuracy, and $\text{U}_{\text{A}}(n_l)=0$ for $n_l$ with minimum accuracy.

The following definitions meet these criteria:

%% Original - Eqn. Version
\vspace{-1em}
\begin{align}
    \text{U}_{\text{L}}(n) &= 1 - \cfrac{L(n)}{\text{max}\,[\, L(n) \,]}\label{eqn:ul}\\
    \text{U}_{\text{A}}(n) &= \cfrac{A(n)}{\text{max}\,[\, A(n) \,]}\label{eqn:ua}
\end{align}

%% Original - Eqn. Version
Now that we have separate utilities for latency and accuracy \ULt{} and \UAt{}, we can combine them to yield an overall utility score.
This allows us to solve the optimization problem by choosing $n$ that maximizes the overall utility.
\cref{eqn:N} represents the solution to this multi-objective optimization problem, defining the overall utility as a convex combination of \ULt{} and \UAt{}.
We measure the effect of different $\alpha$ later in \cref{sec:eval_alpha_ablate}.

\vspace{-1.5em}
\begin{equation}\label{eqn:N}
    R = N - \underset{n}{\text{argmax}}[\; \alpha\text{U}_{\text{A}}(n) + (1-\alpha)\text{U}_{\text{L}}(n) \;]
\end{equation}

\subsection{Pruning Schedule: Deciding Layers at which to Prune}\label{sec:schedule}
We explain \textit{where} in the ViT our pruning mechanism is applied.
Our schedule prunes all $R$ tokens at one layer, early in the model.
This differs from other methods such as ToMe \cite{bolya_token_nodate}, Top-K~\cite{which_tokens_to_use_2023, liang_not_2022}, and DynamicViT~\cite{rao_dynamicvit_2021} that progressively prune tokens.
This choice is based on two observations supported by our evaluation:
First, on small workloads, the repeated application of pruning operations can introduce significant latency (\cref{sec:macrobenchmark}).
Second, latency reductions accumulate with each subsequent layer after pruning; thus, pruning earlier allows more layers to benefit from low latency.

\begin{algorithm}
\begin{minted}[fontsize=\footnotesize]{python}
def vit_forward(model, x, N, R, L):
  """ x is an image-like input. N is 
  the number of tokens after embedding x. 
  R is the number of tokens to prune, and L 
  is the layer at which to prune (Sec. 3.2). """
  # Tokenize input image, x has shape (b, N, d)
  x = model.embed(x)
  for idx, layer in enumerate(model.layers):
    # Standard self-attention
    x, attn, V = layer(x)
    if idx == L:
        scores = rank_tokens(attn, V)
        x = prune_tokens(x, scores, N, R)
  return model.head(x)

def rank_tokens(attn, V):
  """ attn and V are from self-attention.
  attn has shape (b, h, N, N), where h is
  the number of attention heads. V has 
  shape (b, h, N, d / h). 
  'keepdim=True' means the tensor dimension reduced 
  over is not removed, but is kept with length=1.
  """
  am = max(attn,dim=1).sum(dim=1,keepdim=True)
  am /= max(am.transpose(-2,-1))
  # V metric (ours)
  vm = max(V, dim=1).sum(dim=-1,keepdim=True)
  vm = softmax(vm, dim=1)
  # Tokenwise scores with shape (b, N, 1)
  return am + vm 

def prune_tokens(x, scores, N, R):
  """ x is a tensor of tokens. We prune
  such that N-R tokens remain. """
  # Sort by score, descending
  sorted_scores_idx = argsort(scores,dim=1)
  # Shape (b, N-R-1, d)
  kept = gather(x, sorted_scores[:,:N-R-1])
  # Shape (b, 1, d)
  inattentive = gather(x,
    sorted_scores[:,N-R-1:]).mean(dim=1)
  # Shape (b, N-R, d)
  return torch.cat([kept, inattentive], dim=1)
\end{minted}
\caption{Inference-Time Pruning Mechanism}
\label{alg:prune}
\end{algorithm}

We perform pruning after the first 25\% of ViT layers, akin to the first pruning layer of DynamicViT~\cite{rao_dynamicvit_2021} --- this yields latency reduction for the remaining 75\% of layers.
\new{We refer to the index of this pruning layer as $L$.}
Different pruning locations are evaluated in the supplemental work.

%%%
%%% Pruning Method
%%%
\subsection{Token Pruning Method}\label{sec:pruning}
The schedule selection described in~\cref{sec:number_to_prune,sec:schedule} identifies the location and number of tokens to prune.
Now, we give a training-free token pruning mechanism to decide \textit{which} tokens to prune at inference as in \cref{fig:overall_flowchart}.
\new{\textbf{The offline pruning schedule, consisting of a number of tokens to prune $R$ and the layer index $L$ at which to prune, is an input to our token pruning mechanism}.}
The primary design goal for our pruning mechanism is to require no finetuning on \MPre{} and be lightweight - thus we restrict ourselves to using intermediate computations from the attention operation.

%%%
%%% What design choices do we make to satisfy these goals?
%%%
First, we choose a method to rank the importance of tokens.
Following prior work~\cite{bonnaerens_learned_2023, kim_learned_2022}, we rank token importance by measuring the attention each token receives from all others, utilizing the softmax attention matrix.
We also incorporate an importance term derived from the $V$ matrix, which marginally increased accuracy.

Second, we borrow from EViT \cite{liang_not_2022} and instead of discarding pruned tokens, we create a new ``inattentive'' token based on the features of all pruned tokens, then and append it to the set of kept tokens.
Information is thus preserved from the pruned tokens, increasing accuracy while reducing the total token count.

\new{\cref{alg:prune} depicts a forward pass using our pruning mechanism.
Inputs from our offline computation, $R$ and $L$, are utilized to rank tokens and decide at which layer to prune.}
\new{\texttt{rank\_tokens} ranks each token based on our V matrix importance \texttt{vm} and a standard attention matrix importance term \texttt{am}.}
\new{\texttt{prune\_tokens} prunes $R$ tokens using a standard token removal implementation \cite{rao_dynamicvit_2021}.}

%%%
%%% Qualitative Comparison with ToMe
%%%
\subsection{Qualitative Comparison with Pruning and Merging}\label{sec:qual}
\new{
Here, we justify why we classify our method as token pruning rather than merging.
Our method uses the ``inattentive`` token from EViT, which the EViT authors consider a hybrid method.
However, we take the position that the core mechanism of deciding which tokens to prune is an important differentiating factor.
A majority of pruning-based approaches treat the selection of tokens to prune as a ranking problem, rather than a matching problem as merging methods do. 
Our importance score computation is most similar to ranking-based approaches.
Thus, we see ourselves \emph{primarily} as a pruning method, though we could be considered a hybrid between pruning and merging.
}

%%%
%%% Evaluation / Results
%%%
\section{Evaluation}\label{sec:eval}
After describing experimental setup (\cref{sec:experimental_setup}),
we characterize our technique via ablation over $\alpha$ \new{and by measuring offline computation costs (\cref{sec:microbenchmark})}.
  Then we compare to the state-of-art ToMe method and relevant baselines (\cref{sec:macrobenchmark}).

\subsection{Experimental Setup}\label{sec:experimental_setup}

\textbf{Hardware}: We use three devices with varying characteristics (\cref{tab:hardware_info}).
Our technique targets edge workloads, so we use two edge-caliber development boards designed for machine learning:
NVIDIA TX2~\cite{nvidia_tx2} and NVIDIA AGX Orin~\cite{nvidia_agxorin}.
To assess generalization beyond edge devices, we also use a server-grade NVIDIA A100 GPU~\cite{nvidia_a100}.
On the TX2 and Orin we used fixed CPU and GPU clock rates for consistency.
The A100 system clocks could not be locked because those servers are shared resources.

\begin{table}[!h]
    \centering
    \setlength{\tabcolsep}{8pt}
    \begin{tabular}{lrrrr}
        \toprule
        %Device Name & VRAM (GB) & CPU Cores & GPU Cores & CUDA Cores & Max Power (W)\\
        \textbf{Device}  & \textbf{GPU} & \textbf{CUDA} & \textbf{Max Power}\\
        & \textbf{Cores} & \textbf{Cores} & \textbf{(W)}\\
        \toprule
        %TX2 & 4 & 4 & 1 & 256 & 15\\
        %AGX Orin & 8 & 12 & 2 & 2048 & 60\\
        %A100 & 80 & 32 & 128 & 8192 & 400*\\
        TX2 \cite{nvidia_tx2} & 2 & 256 & 15\\
        Orin \cite{nvidia_agxorin} & 14 & 1792 & 40\\
        A100 \cite{nvidia_a100} ~\ding{61} & 108 & 6192 & 300\\
        \bottomrule
    \end{tabular}
    \caption{
    Summary of device hardware information. We evaluate our method on two edge devices (TX2 and AGX Orin) and one server-grade system in this work. 
    ~\ding{61} The A100 power consumption is the power consumption only for the GPU, not the entire system.
    }
    \label{tab:hardware_info}
\end{table}%
\begin{table}[!t]
    \centering
    \setlength{\tabcolsep}{8pt}
    \begin{tabular}{lrrr}
        \toprule
        \textbf{Model}  & \textbf{Params (M)} & \textbf{Depth}\\
        \toprule
        DeiT-S~\cite{touvron_deit_2022} & 21 & 12\\
        DeiT-B~\cite{touvron_deit_2022} & 86 & 12\\
        ViT-L~\cite{dosovitskiy_image_2020,huggingface_timm} & 300 & 24\\
        DinoV2-G~\cite{oquab_dinov2_2023} & 1100 & 40\\
        \bottomrule
    \end{tabular}
    \caption{
    Characteristics of models we evaluate over.
    We include DinoV2 as a representative state-of-the-art ViT, while ViT and DeiT are commonly used baselines in prior token pruning work.
    }
    \label{tab:model_info}
\end{table}

\textbf{Models}:
\new{\cref{tab:model_info} summarizes the models used.}
We evaluate common vision transformer models across a variety of scales (21M to 1.1B parameters).
For DeiT and ViT models we use the TIMM pretrained weights, while we use the DinoV2-G weights from the DinoV2 Github.

\textbf{Measurements}:
In order to measure latency for evaluation, we use the PyTorch \texttt{benchmark} module~\cite{torch_benchmark}.
\new{Latency is measured over 16 seconds.}
\new{\textbf{To be clear, we define latency as the compute time required for a forward pass of a model given a batch of input images}.}
Accuracy was measured using the A100 system, with a batch size of 512 on the classification evaluation subset of ImageNet1K.
In subsequent experiments, we decide the pruning hyperparameters of each token pruning method for fair comparison with our method.
For reproducibility, our code is open source. 

\subsection{Characterizations of our technique}\label{sec:microbenchmark}
Here we evaluate \new{two of the three} design decisions of our method.
First, we ablate the hyperparameter $\alpha$ used in our utility function.
Second, we measure the cost (time) for offline computation.
\new{The third decision, selecting the layer $L$ for pruning, is evaluated in supplemental material.}

%%%
%%% Alpha ablation study
%%%
%\subsubsection{Pruning schedule ablation study across $\alpha$}\label{sec:eval_alpha_ablate}
\textbf{Pruning schedule ablation study across $\alpha$. }\label{sec:eval_alpha_ablate}
In~\cref{sec:number_to_prune} we introduce an algorithm to decide a number of tokens to prune $R$ according to the GPU tail effect.
%%%
%%% Alpha Table Ablation
%%%
%%% PURPOSE ? 
%%% Have a table that shows when token pruning can actually DEGRADE latency!
%%% Includes our method, but also shows some instances where 
\begin{table}[!htb]
    \centering
    \setlength{\tabcolsep}{7pt}
    \begin{tabular}{lllrr}
        \toprule
        \textbf{Device} & \textbf{$\alpha$} & \textbf{R} & $\downarrow$\textbf{Top-1} & $\downarrow$\textbf{Median}\\
        ~ & \textbf{Range} & ~ & \textbf{Loss} & \textbf{Latency (ms)}\\
        \toprule
        \multirow{3}{*}{Orin} & $[0.1, \;0.3]$ & 139 & 1.07 & \textcolor{blue}{(-27.8\%)} 112.2\\
        & $[0.4, \;0.5]$ & 166 & 2.04 & \textcolor{blue}{(-33.0\%)} 104.2\\
        & $[0.6, \;0.9]$ & 193 & 4.82 & \textcolor{blue}{(-35.1\%)} 100.9\\
        \midrule
        \multirow{2}{*}{A100} & $[0.1, \;0.4]$ & 73 & 0.25 & \textcolor{blue}{(-4.70\%)} 28.77\\
        & $[0.5, \;0.9]$ & 139 & 1.07 & \textcolor{blue}{(-6.34\%)} 28.27\\
        \bottomrule
    \end{tabular}
    \caption{
    Accuracy vs latency tradeoffs for the computed number of tokens to prune $R$ for various $\alpha$ of our method.
    Latency percent change is with respect to baseline DinoV2-G inference time.
    %Note that latency increased as \textit{more} tokens were removed in the AGX Orin trial.
    %Latency can be variable for small workload sizes.
    }
    \label{tab:alpha_ablate}
\end{table}
%The hyperparameter of this algorithm is $\alpha$, which determines the relative weighting of the accuracy and latency utilities.
The $\alpha$ hyperparameter governs the relative weighting of accuracy and latency utilities in our algorithm.
Setting $\alpha=1$ prioritizes accuracy, and only a few tokens might be pruned.
Setting $\alpha=0$ prioritizes latency, and would prune all or nearly-all tokens.
To decide the value of $\alpha$, we performed an ablation study in ~\cref{tab:alpha_ablate}.

%Evaluation on the AGX Orin for $\alpha > 0.5$ yielded $R=193$ (over 75\% of tokens pruned), which degraded accuracy by $\sim$2.8\% for 2.1\% less latency.
When evaluated on the AGX Orin with $\alpha > 0.5$, we computed $R=193$ (over 75\% of tokens pruned), resulting in a $\sim$2.8\% accuracy drop for a 2.1\% latency reduction.
We consider this to be an unfavorable tradeoff.
This ablation, in addition to results in \cref{sec:eval_tradeoffs}, suggest that $\alpha \leq 0.5$ is a good choice.
Thus we use $\alpha=0.5$ for all evaluations that appear in this work.
Intuitively speaking this means accuracy degradation and latency reduction are considered with equal weight according to \cref{eqn:N} when selecting $R$.

%%%
%%% Grid Search Time
%%%
%\subsubsection{Offline Computation Time}\label{sec:offline_time}
\textbf{Offline computation time. }\label{sec:offline_time}
\new{
In ~\cref{sec:number_to_prune}, we describe our offline computations to decide a \textit{number} of tokens to prune in which we utilize a grid search to measure latency and estimate accuracy degradation.
~\cref{tab:offline_compute_time} illustrates the total times required for measuring the workload-latency characteristics of DeiT-S and ViT-L across devices.
The offline computation is relatively fast (no more than 4.5 hours), especially compared with the time required to train any ViT.
}
\begin{table}[!hbt]
    \centering
    \setlength{\tabcolsep}{8pt}
    \setlength{\belowcaptionskip}{-6pt}
    \begin{tabular}{llrr}
        \toprule
        \textbf{Device} & \textbf{Model} & \textbf{$\dagger$Accuracy} & \textbf{Latency} \\
        ~ & ~ & \textbf{Time (min)} & \textbf{Time (min)} \\
        \toprule
        \multirow{2}{*}{TX2} & DeiT-S & - & 54 \\
        ~ & ViT-L & - & 54 \\
        \midrule
        %%%
        \multirow{2}{*}{Orin} & DeiT-S & - & 54 \\
        ~ & ViT-L & - & 54 \\
        \midrule
        %%%
        \multirow{2}{*}{A100} & DeiT-S & 83 & 54 \\
        ~ & ViT-L & 218 & 52 \\
        \bottomrule
    \end{tabular}
    \caption{
    Time to estimate accuracy degradation and measure latency as described in \cref{sec:number_to_prune} for our offline computation.
    $\dagger$Accuracy was measured on the A100 with, since it does not depend on latency characteristics of the target device.
    %Time refers to the number of minutes to measure the workload-latency relationship on each device and model.
    %Stride is the granularity of the grid-search; stride=1 means we evaluate every possible number of tokens, while stride=2 means we evaluate every other token, etc.
    }
    \label{tab:offline_compute_time}
\end{table}

%%%
%%% Similar Latency Data Cross-Device
%%%
% \input{tables/table_similar_latency}
\begin{table}[!hbt]
    \centering
    \setlength{\tabcolsep}{3.5pt}
    \setlength{\belowcaptionskip}{-8pt}
    %%% Device, Model + Wrapper, Accuracy, Latency
    \begin{tabular}{llllrrr}
        \toprule
        \textbf{Device} & \textbf{Batch} & \textbf{Model} & ~ & $\downarrow$\textbf{Top-1} & $\downarrow$\textbf{Median}\\
        ~ & \textbf{Size} & ~ & ~ & \textbf{Loss} & \textbf{Latency (ms)}\\
        \toprule
        %%%
        %%% Jetson TX2
        %%%
        %%% DeiT-S
        \multirow{5}{*}{TX2} & \multirow{5}{*}{2} & DeiT-S & ~ & ~ & \phantom{-(00.0\%)} 68.92\\
        & & w/ Top-K & ~ & 4.30 & \textcolor{blue}{(-27.0\%)} 50.32\\
        & & w/ ToMe  & ~ & 2.47 & \textcolor{blue}{(-23.1\%)} 52.98\\
        & & w/ DyViT & ~ & \ding{61} 0.46  & \textcolor{blue}{(-26.2\%)} 50.88\\
        & & w/ Ours  & ~ & 1.24 & \textcolor{blue}{(-28.3\%)} 49.44\\
        \midrule
        %%% DeiT-B
        \multirow{5}{*}{TX2} & \multirow{5}{*}{2} & DeiT-B & ~ & ~ & \phantom{-(00.0\%)} 215.0\\
        & & w/ Top-K & ~ & 2.44 & \textcolor{blue}{(-33.5\%)} 143.0\\
        & & w/ ToMe  & ~ & 1.63 & \textcolor{blue}{(-33.7\%)} 142.7\\
        & & w/ DyViT & ~ & \ding{61} 0.57 & \textcolor{blue}{(-33.2\%)} 143.7\\
        & & w/ Ours  & ~ & 1.16 & \textcolor{blue}{(-32.1\%)} 146.0\\
        \midrule
        %%% VIT-L
        %%% Baseline Acc 84.378
        %%% Top-K Acc 46.184
        %%% ToMe Acc 66.874
        %%% HATP Acc 75.944
        \multirow{4}{*}{TX2} & \multirow{4}{*}{2} & ViT-L & ~ & ~ & \phantom{-(00.0\%)} 1327.0\\
        & & w/ Top-K & ~ & 38.2 & \textcolor{blue}{(-57.2\%)} 568.0\\
        & & w/ ToMe  & ~ & 17.5 & \textcolor{blue}{(-57.8\%)} 559.3\\
        & & w/ Ours  & ~ & 8.4 & \textcolor{blue}{(-55.2\%)} 594.9\\
        \midrule
        %%%
        %%% AGX Orin
        %%%
        %%% DinoV2-G
        %%% VIT-L
        %%% Top-K R=10, Acc= 31.674
        %%% ToMe R=15, Acc=66.864
        %%% ViT-L baseline accuracy is 84.378
        \multirow{4}{*}{Orin} & \multirow{4}{*}{4} & ViT-L & ~ & ~ & \phantom{-(00.0\%)} 70.29\\
        & & w/ Top-K & ~ & 52.70 & \textcolor{blue}{(-32.2\%)} 47.63\\
        & & w/ ToMe  & ~ & 17.51 & \textcolor{blue}{(-22.9\%)} 54.18\\
        & & w/ Ours  & ~ & 2.35 & \textcolor{blue}{(-33.0\%)} 47.08\\
        \midrule
        \multirow{4}{*}{Orin} & \multirow{4}{*}{2} & DinoV2-G & ~ & ~ & \phantom{-(00.0\%)} 155.5\\
        & & w/ Top-K & ~ & 45.66 & \textcolor{blue}{(-32.9\%)} 104.4\\
        & & w/ ToMe & ~ & 6.96 & \textcolor{blue}{(-32.9\%)} 104.4\\
        & & w/ Ours & ~ & 2.04 & \textcolor{blue}{(-33.1\%)} 104.1\\
        \midrule
        %%%
        %%% A100
        %%%
        %%% ViT-L
        %%% ToMe and Top-K have R=1
        % \multirow{4}{*}{A100} & \multirow{4}{*}{16} & ViT-L & ~ & ~ & \phantom{-(00.0\%)} 29.41\\
        % & & w/ Top-K & ~ & 0 & \textcolor{red}{(+2.48\%)} 30.14\\
        % & & w/ ToMe  & ~ & 0 & \textcolor{red}{(+7.65\%)} 31.66\\
        % & & w/ Ours  & ~ & 0 & \textcolor{blue}{(-26.3\%)} 21.68\\
        % \midrule
        %%% DinoV2-G
        \multirow{4}{*}{A100} & \multirow{4}{*}{4}  & DinoV2-G & ~ & ~ & \phantom{-(00.0\%)} 40.53\\
        & & w/ Top-K & ~ & 13.31 & \textcolor{blue}{(-16.7\%)} 33.76 \\
        & & w/ ToMe & ~ & 6.96 & \textcolor{blue}{(-16.6\%)} 33.79 \\
        & & w/ Ours & ~ & 2.04 & \textcolor{blue}{(-20.1\%)} 32.37 \\
        \bottomrule
    \end{tabular}
    \caption{
    Comparison with existing methods for similar latency across devices and workload sizes.
    Hyperparameters of methods were chosen to get as close as possible to match our latency; \new{in some cases pruning more tokens did not reduce latency.}
    \new{\textbf{For each model, the smallest batch size was selected to demonstrate cases where the workload-latency relationship can be exploited.}}
    \ding{61}DynamicViT (DyViT) is the the only training-based method.
    }
    \label{tab:similar_latency}
\end{table}

%%%
%%% Pareto Scatter Plot Figures
%%%
\begin{figure}[!t]
\centering
\setlength{\belowcaptionskip}{-6pt}
\includegraphics[width=0.80\linewidth]{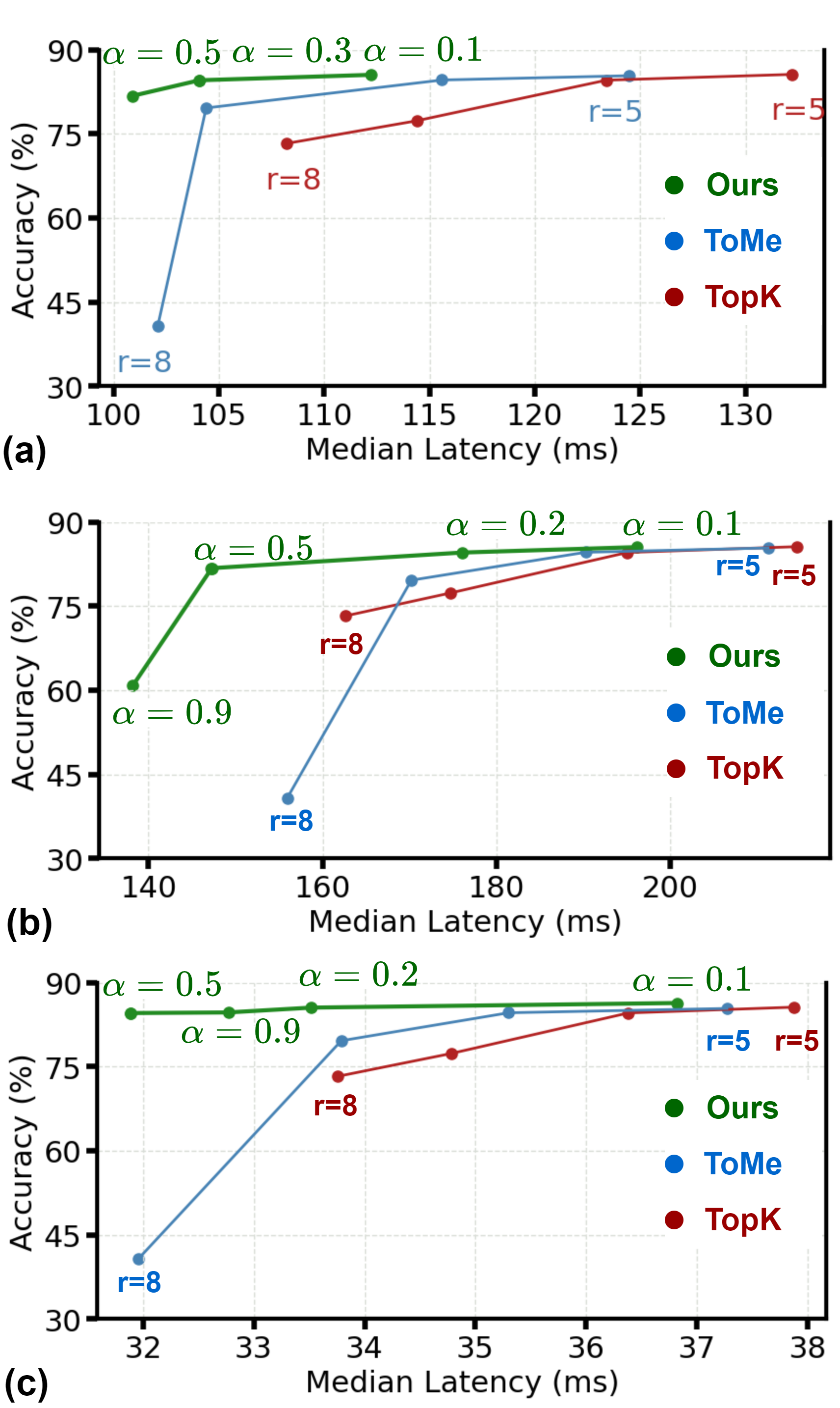}
\caption{
Illustration of accuracy-latency tradeoffs of surveyed methods with \MPre{} = DinoV2-G: (a) batch size=2 on AGX Orin (b) batch size=4 on A100 (c) batch size=4 on AGX Orin.
Our pruning schedule and mechanism generate points that expand the pareto front. 
The number of tokens removed at each layer ($r$) of Top-K and ToMe is evaluated from $r=5$ to $=8$ in increments of 1.
}
\label{fig:pareto}
\end{figure}

\subsection{Comparison to Other Methods}\label{sec:macrobenchmark}
In this section, we demonstrate the effectiveness of our token pruning schedule and pruning mechanism across devices and workload sizes.
Our primary focus is on smaller workloads, which are typical in edge deployment scenarios~\cite{edge_optim_batch_2022}.

\new{
For inter-method comparison, we systematically compare to the state-of-the-art method, Token Merging (ToMe)~\cite{bolya_token_nodate}.
We also evaluate two common benchmarks, Top-K~\cite{which_tokens_to_use_2023, liang_not_2022} and DynamicViT~\cite{rao_dynamicvit_2021}.
We measure Top-K in all conditions; we use DynamicViT only with models for which its pre-trained weights were available.
DynamicViT is not emphasized as it involves training, making it an unfair comparison with our method, ToMe, and Top-K.
}

%%%
%%% Accuracy Latency Tradeoffs
%%%
\textbf{Accuracy-latency tradeoffs. }\label{sec:eval_tradeoffs}
In this section, we discuss the results of our training-free token pruning mechanism and offline computed pruning schedule.
\cref{tab:similar_latency} illustrates our method's ability to retain higher accuracy for similar latency across devices and workload sizes.
\cref{fig:pareto} demonstrates that our token pruning mechanism and schedule expands the accuracy-latency tradeoffs on the pareto front across devices and workload sizes.

%%%
%%% Similar Latency Talk
%%%
First, our method is able to achieve higher accuracy than other training-free pruning techniques ToMe and Top-K.
\cref{tab:similar_latency} is an ablation study across various workload sizes (batch size, models) and devices, where we tune the hyperparameters of methods such that similar latency is achieved.
Unsurprisingly, DynamicViT retains accuracy since it was finetuned for 300 epochs.
However, compared to ToMe and Top-K, our method consistently results in lower accuracy degradation.
In one case, Top-K degrades accuracy by more than 45.6\%, while we degrade accuracy by only 2\%.
Across all workload sizes and devices, ToMe had 0.47 to 15.16 \textit{lower} Top-1 percentage points than our method for similar latency (within 5.2\% or 7ms).

%%%
%%% Pareto Talk
%%%
Second, we note that for larger pruning rates such as $r=7$ and $r=8$, ToMe and Top-K remove nearly all input tokens by the last layer of DinoV2-G.
%as seen in \cref{fig:pareto}.
These high pruning ratios yield high accuracy degradation of over 40\% in the case of A100 batch-size 4 for ToMe, as seen in \cref{fig:pareto}.
Comparatively, by pruning 54\%-75\% of tokens at the 10th layer of DinoV2-G according to the tail effect, we achieve higher accuracy and lower latency.

In both \cref{tab:similar_latency} and \cref{fig:pareto}, our method features lower accuracy degradation than ToMe and Top-K.
At small workloads, the marginal latency of pruning additional tokens becomes negligible.
As a result, \textit{pruning tokens at each layer degrades accuracy significantly for little latency benefit; existing methods do not account for this behavior.}
Thus, we prune $R$ tokens early; the remaining tokens propagate through the ViT, retaining information which leads to better accuracy.
Simultaneously, we achieve high latency reduction due to pruning early in the network.

%%%
%%% Workload Size
%%%
\textbf{Low workload size observations. }\label{sec:eval_small_workload}
For small workloads, token sparsification can actually increase latency due to the overhead associated with token removal mechanisms.
\cref{tab:latency_anomaly} illustrates three examples of this.
\new{ToMe and Top-K may \textit{increase} latency by 2-30\% with respect to baseline, while we \textit{reduce} it by 9-26\%.
There are also cases where all methods, including ours, increase latency by 40\%-134\%.}
We find that for some workloads, using a baseline model or our method is strictly better than attempting to use a token removal method with high overhead.
%%%
%%% Latency Anomaly Table
%%%
%\input{tables/table_anomaly_latency}
%%% PURPOSE ? 
%%% Have a table that shows when token pruning can actually DEGRADE latency!
%%% Includes our method, but also shows some instances where 
\begin{table}[!t]
    \centering
    \setlength{\tabcolsep}{4pt}
    \setlength{\belowcaptionskip}{-6pt}
    %%% Device, Model + Wrapper, Accuracy, Latency
    \begin{tabular}{lllrr}
        \toprule
        \textbf{Device} & \textbf{Batch} & \textbf{Model} & $\downarrow$\textbf{Top-1} & $\downarrow$\textbf{Median}\\
        ~ & \textbf{Size} & & \textbf{Loss} & \textbf{Latency (ms)}\\
        \toprule
        %%%% TX2
        %%% Top-K r=6
        %%% ToMe r=6
        %%% Ours r=69
        \multirow{4}{*}{TX2} & \multirow{4}{*}{1} & DeiT-S & ~ & \phantom{(-0.00\%)} 35.32\\ 
        & & Top-K & 0.73 & \textcolor{red}{(+18.6\%)} 43.38\\
        & & ToMe & 0.27 & \textcolor{red}{(+30.3\%)} 50.70\\
        & & Ours & 0.99 & \textcolor{blue}{(-9.06\%)} 32.12\\
        \midrule
        %%%% A100
        %%% Top-K r=3
        %%% ToMe r=3
        %%% Ours r=69
        %%% ViT-L baseline accuracy is 84.378
        \multirow{4}{*}{A100} & \multirow{4}{*}{4} & ViT-L & ~ & \phantom{(-0.00\%)} 9.49\\ 
        & & Top-K & 0.47 & \textcolor{red}{(+51.4\%)} 14.36\\
        & & ToMe & 0.29 & \textcolor{red}{(+134.4\%)} 22.24\\
        & & Ours & 0.85 & \textcolor{red}{(+40.0\%)} 13.29\\
        \midrule
        %%% A100
        \multirow{4}{*}{A100} & \multirow{4}{*}{16} & ViT-L & ~ & \phantom{-(00.0\%)} 29.41\\
        & & w/ Top-K & 0.77 & \textcolor{red}{(+2.48\%)} 30.14\\
        & & w/ ToMe & 0.51 & \textcolor{red}{(+7.65\%)} 31.66\\
        & & w/ Ours & 2.26 & \textcolor{blue}{(-26.3\%)} 21.68\\
        \bottomrule
    \end{tabular}
    \caption{
    Experiment that illustrate pruning overhead for certain workload sizes.
    \new{Pruning parameters were chosen such that a similar number of tokens were pruned as our method.}
    Token pruning methods may \textit{increase} latency due to the overhead of pruning itself.
    We reduce overhead through single-layer pruning.
    }
    \label{tab:latency_anomaly}
\end{table}

%%%
%%% Limitations
%%%
\section{Limitations}\label{sec:limitations}
Our token pruning approach is optimized for cases where latency-workload relationships are non-linear.
%This occurs primarily for small workload and batch sizes, which are common contexts in edge deployment.
For large workloads such as DinoV2-G with batch size 256+, our method is less effective because latency-workload relationships becomes linear and more predictable in this case.

%%%
%%% Conclusion
%%%
\section{Conclusion}\label{sec:conclusion}
In this work, we offered practical guidance on how to improve token pruning for ViTs in the presence of small workloads by utilizing latency-workload relationships.
%
%First, we provided insight on how ViT latency is affected by underlying hardware, machine learning framework, and kernel launch factors.
%
%Second,
We showed how to determine a token pruning schedule by leveraging non-linear latency-workload relationships; in comparison with prior work, our method yields equal or greater latency reductions while maintaining greater accuracy.
%
%Third, we introduced a training-free, low-overhead token pruning mechanism.
%
%Our method advanced the accuracy-latency Pareto front compared with existing token pruning approaches across small workload sizes and devices.
%
Ultimately, we demonstrated that leveraging workload-latency behavior is effective at improving ViT efficiency via token pruning, especially for small workloads.
%
%We show that for similar latency (within 5.2\% or 7ms) across devices, the state-of-the-art, ToMe, achieves 45.8\%-85.4\% ImageNet1K accuracy, while we achieve 78.6\%-84.5\%.

%%%
%%% References
%%%
\clearpage
{\small
\bibliographystyle{ieee_fullname}
\bibliography{references}
}

%%%
%%% Supplemental
%%%
\ifSUPPLEMENTAL
\clearpage

\section{Supplemental Material}
This supplemental data presents details from \cref{tab:similar_latency} in \cref{sec:eval_tradeoffs}, and explaining limitations of our method mentioned in \cref{sec:limitations}.
First, in \cref{sec:eval_prune_loc_ablate} we ablate over various pruning locations for our method.
Second, in \cref{sec:prune_hyperparam} we list the hyperparameters of Top-K and ToMe pruning methods used in evaluation with our method, in case others want to reproduce our work.
Third, in \cref{sec:large_tradeoff} we show an example in which the accuracy-latency tradeoffs of our method become less significant at larger workload sizes.

%%%
%%% Small Table of Various Pruning Locations
%%%
\subsection{Pruning location ablation study}\label{sec:eval_prune_loc_ablate}
%\NJE{Backup Plan: Move this to appendix to make room since it is of relatively low importance wrt. the other data}
In~\ref{sec:schedule} we decide at which layer our pruning mechanism should be applied.
% In order 
To provide insight into  the potential pruning locations, we performed an ablation study.
\cref{tab:location_ablate} illustrates latency and accuracy tradeoffs for various pruning locations of DinoV2-G.

\begin{table}[!ht]
    \setlength{\belowcaptionskip}{-6pt}
    \setlength{\tabcolsep}{12pt}
    \centering
    \begin{tabular}{llrr}
    \toprule
    \textbf{Batch} & \textbf{Pruning} & $\downarrow$\textbf{Acc.} & $\downarrow$\textbf{Median}\\
    \textbf{Size} & \textbf{Layer} & \textbf{Loss} & \textbf{Latency (ms)}\\
    \toprule
    %%% Batch Size 1
    \multirow{4}{*}{1} & \phantom{0}1/40 & 3.19 & 68.4\\
    & 10/40 & 1.07 & 81.3\\
    & 20/40 & 0.58 & 93.4\\
    & 30/40 & 0.49 & 104.4\\
    \midrule
    %%% Batch Size 2
    \multirow{4}{*}{2} & \phantom{0}1/40 & 7.08 & 79.1\\
    & 10/40 & 2.04 & 104.7\\
    & 20/40 & 0.97 & 133.0\\
    & 30/40 & 0.83 & 160.8\\
    \bottomrule
    \end{tabular}
    \caption{
    Latency/accuracy tradeoffs by pruning location. Configuration: \MPre{} = DinoV2-G on AGX Orin.
    }
    \label{tab:location_ablate}
\end{table}

As expected, pruning earlier yields lower latency but greater accuracy degradation.
For the batch size 1, our method pruned $\sim$54\% of input tokens at the first layer degraded accuracy by 3.19\% but yielded a 40\% overall latency reduction.
Across both batch size 1 and 2 in this ablation study, pruning after the first 25\% of layers (layer 10) results in a good balance between latency reduction and accuracy degradation.

Pruning later in the network will reduce accuracy degradation, however we prioritize yielding latency benefits with our method.
Therefore, we perform pruning at the layer 25\% of the way into the network for all models evaluated in this work, as stated in \cref{sec:schedule}.

%%%
%%% Batch Size Graphic (Shrunk hopefully)
%%%
\begin{table}[!ht]
    \centering
    \setlength{\tabcolsep}{4pt}
    \setlength{\belowcaptionskip}{-8pt}
    %%% Device, Model + Wrapper, Accuracy, Latency
    \begin{tabular}{lllrrr}
        \toprule
        \textbf{Device} & \textbf{Batch} & \textbf{Model \&} & \textbf{Acc.} & \textbf{Median} \\
        ~ & \textbf{Size} & \textbf{Method} & \textbf{Loss} & \textbf{Latency (ms)}\\
        \toprule
        %%%
        %%% Jetson TX2
        %%%
        %%% Batch Size 2
        %%% DeiT-B
         \multirow{5}{*}{TX2} & \multirow{5}{*}{2} & Top-K $r$=13 & 2.44 & \textcolor{blue}{(-33.5\%)} 143.0\\
         & & ToMe $r$=12 & 1.63 & \textcolor{blue}{(-33.7\%)} 142.7\\
         & & Ours $R$=70 & 1.16 & \textcolor{blue}{(-32.1\%)} 146.0\\
         & & DyViT $K$=0.68 & \ding{61} 0.57 & \textcolor{blue}{(-33.2\%)} 143.7\\
         & & \textbf{DeiT-B} & ~ & \phantom{-(00.0\%)} 215.0\\
         \midrule
        %%% DeiT-S
         \multirow{5}{*}{TX2} & \multirow{5}{*}{2} & Top-K $r$=15 & 4.30 & \textcolor{blue}{(-27.0\%)} 50.32\\
         & & ToMe $r$=17 & 2.47 & \textcolor{blue}{(-23.1\%)} 52.98\\
         & & Ours $R$=77 & 1.24 & \textcolor{blue}{(-28.3\%)} 49.44\\
         & & DyViT $K$=0.70 & \ding{61} 0.46  & \textcolor{blue}{(-26.2\%)} 50.88\\
         & & \textbf{DeiT-S} & ~ & \phantom{-(00.0\%)} 68.92\\
        \midrule
        %%%
        %%% AGX Orin
        %%%
        %%% Batch Size 2
        %%% DinoV2-G
        \multirow{4}{*}{Orin} & \multirow{4}{*}{2} & Top-K $r$=9 & 45.66 & \textcolor{blue}{(-32.9\%)} 104.4\\
         & & ToMe $r$=7 & 6.96 & \textcolor{blue}{(-32.9\%)} 104.4\\
         & & Ours $R$=166 & 2.04 & \textcolor{blue}{(-33.1\%)} 104.1\\
         & & \textbf{DinoV2-G} & ~ & \phantom{-(00.0\%)} 155.5\\
        \midrule
        %%%
        %%% A100
        %%%
        %%% Batch Size 2
        %%% DinoV2-G
        \multirow{4}{*}{A100} & \multirow{4}{*}{4} & Top-K $r$=8 & 13.31 & \textcolor{blue}{(-16.7\%)} 33.76\\
         & & ToMe $r$=7 & 6.96 & \textcolor{blue}{(-16.6\%)} 33.79\\
         & & Ours $R$=166  & 2.04 & \textcolor{blue}{(-20.1\%)} 32.37\\
         & & \textbf{DinoV2-G} & ~ & \phantom{-(00.0\%)} 40.53\\
        \bottomrule
    \end{tabular}
    \caption{
    Companion table to \cref{tab:similar_latency} with hyperparameters annotated for each entry of the original table.
    Top-K \cite{which_tokens_to_use_2023} and ToMe \cite{bolya_token_nodate} remove $r$ tokens each layer. 
    $R$ refers to the number of tokens our method decides to prune, and $K$ is DynamicViT's keep ratio \cite{rao_dynamicvit_2021} (it is also referred to as $r$ in their work, however we redefine it since it is used differently than the $r$ of ToMe and Top-K).
    \ding{61} Note that DynamicViT requires training, while all other evaluated methods are training free.
    }
    \label{tab:similar_latency_hyperparams}
\end{table}
\begin{figure}[!tbh]
    \setlength{\belowcaptionskip}{-8pt}
    \centering
    \includegraphics[width=0.74\linewidth]{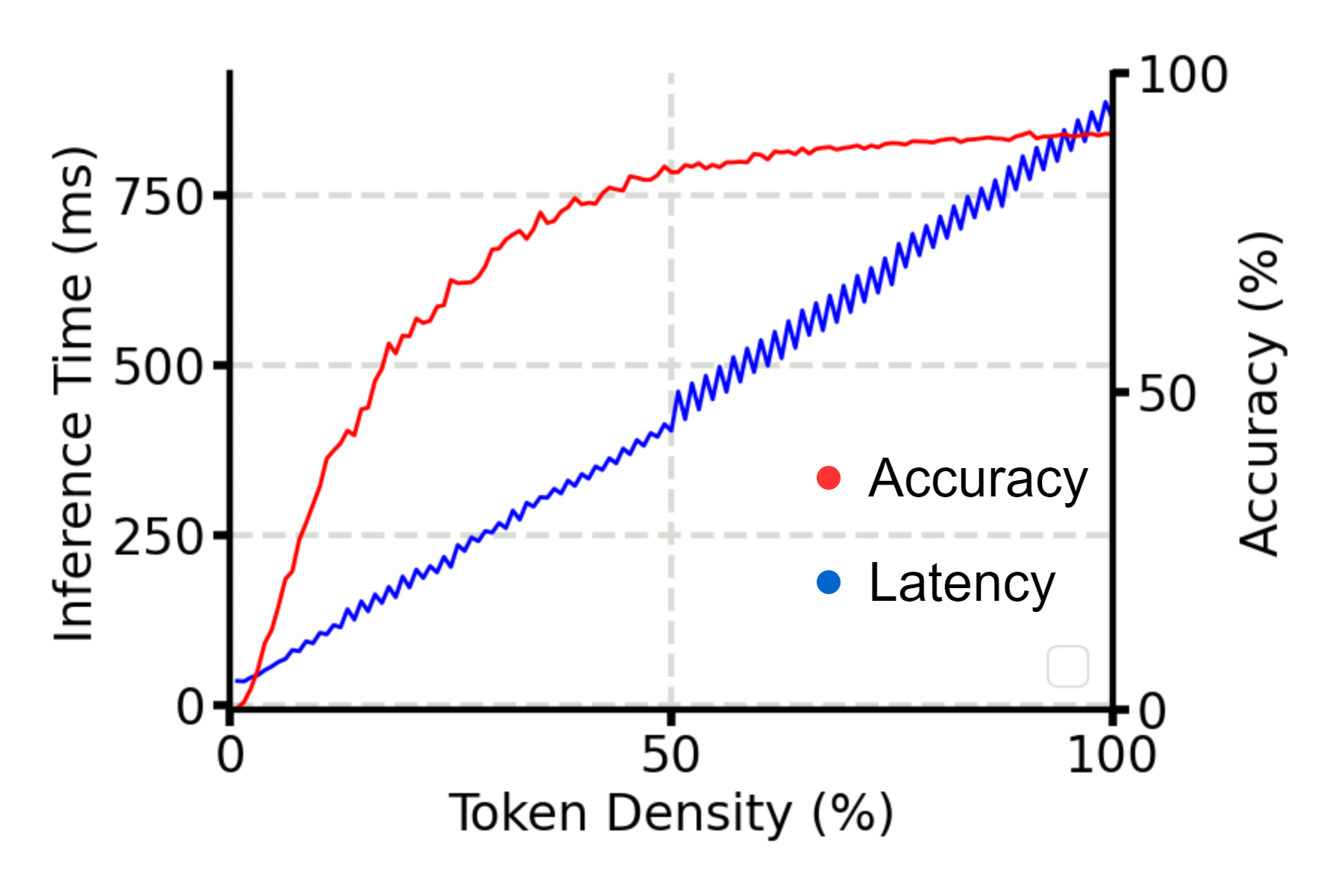}
    \caption{GPU Tail Effect has less impact on large batch size (here, the AGX Orin on DeIT-B with batch size of 128).}
    \label{fig:batch_128}
\end{figure}

%%%
%%% Hyperparameters from Experiments
%%%
\subsection{Pruning Hyperparameters Used for Comparison with Other Work}\label{sec:prune_hyperparam}
In \cref{tab:similar_latency} we perform an experiment where we show the differences in accuracy of our method and others across models and devices.
In \cref{tab:similar_latency_hyperparams} we present the same data annotated with an extra column for the hyperparameters of each method.
Note that in both tables hyperparameters are chosen such that all methods achieve similar latency to our method.

\subsection{Large Workload Size Tradeoffs}\label{sec:large_tradeoff}
In \cref{sec:limitations}, we hypothesize that our method may achieve worse tradeoffs for larger workload sizes.
Our method prioritizes pruning a number of tokens for which large latency changes occur.
However, at larger workload sizes the latency-workload relationship becomes more linear.
\cref{fig:batch_128} depicts this phenomena for DeiT-B on the AGX Orin with batch size 128.
It can be seen there are no large changes in latency to exploit, which is how our method is able to outperform other techniques like ToMe for small workload sizes.

\else\fi
\end{document}